\documentclass[final,5p,times,twocolumn]{elsarticle}

\usepackage[colorlinks,
 linkcolor=blue,
 anchorcolor=blue,
 citecolor=blue
 ]{hyperref}
\usepackage{amsthm}
\usepackage{amsmath}
\usepackage{stfloats}
\biboptions{sort&compress}
\usepackage{booktabs}
\usepackage{url}

\usepackage{amssymb}
\usepackage{graphicx}
\usepackage{cleveref}
\usepackage{picins}

\journal{Neurocomputing}
\begin{document}

\begin{frontmatter}



\title{Dynamically Retrieving Knowledge via Query Generation for Informative Dialog Generation}
\address[1]{Northwestern Polytechnical University,China}
\address[2]{Xidian University,China}
\author[1]{Zhongtian Hu}
\ead{ahxchzt@mail.nwpu.edu.cn}
\author[1]{Lifang Wang}
\author[1]{Yangqi Chen}
\author[1]{Yushuang Liu}
\author[2]{Ronghan Li}
\author[1]{Meng Zhao}
\author[1]{Xinyu Lu}
\author[1]{Zejun Jiang\corref{cor1}}
\ead{claud@nwpu.edu.cn}

\cortext[cor1]{Corresponding author}

\begin{abstract}
	Knowledge-driven dialog system has recently made remarkable breakthroughs. Compared with general dialog systems, superior knowledge-driven dialog systems can generate more informative and knowledgeable responses with pre-provided knowledge. However, in practical applications, the dialog system cannot be provided with corresponding knowledge in advance because it cannot know in advance the development of the conversation. Therefore, in order to make the knowledge dialogue system more practical, it is vital to find a way to retrieve relevant knowledge based on the dialogue history. To solve this problem, we design a knowledge-driven dialog system named DRKQG (Dynamically Retrieving Knowledge via Query Generation for informative dialog response). Specifically, the system can be divided into two modules: the query generation module and the dialog generation module. First, a time-aware mechanism is utilized to capture context information, and a query can be generated for retrieving knowledge through search engine. Then, we integrate the copy mechanism and transformers, which allows the response generation module to produce responses derived from the context and retrieved knowledge. Experimental results at LIC2022, Language and Intelligence Technology Competition, show that our module outperforms the baseline model by a large margin on automatic evaluation metrics, while human evaluation by the Baidu Linguistics team shows that our system achieves impressive results in Factually Correct and Knowledgeable.
\end{abstract}

%

\begin{keyword}
	Response generation \sep
	Knowledge-driven dialogue system \sep
	Query generation



\end{keyword}

\end{frontmatter}



\section{Introduction}\label{Indro}
Response generation has always been a significant challenge in dialog systems, and related research has rapidly evolved from the seq2seq model \cite{sutskever195sequence} to pre-trained language models (PrLMs). Various approaches have been proposed \cite{wu2018neural,li2016diversity,2019Gating,2019A}, which can generate fluent responses. However, these approaches only utilize utterances from the dialog history. They tend to generate uninformative and dull responses such as “I am fine” and “You are right,” which severely spoil the user experience. Thus, knowledge-based dialog systems \cite{moghe2018towards,zhou2020kdconv,moon2019opendialkg,2019Knowledge,2021Towards} have recently received extensive attention because these dialog systems can utilize the provided structured or unstructured knowledge to produce more knowledgeable responses.

Most research on knowledge-driven dialog systems has focused on knowledge injection tasks. For example, some works use a complex attention mechanism to fuse knowledge and context \cite{lin2020generating}, and several works perform knowledge selection task before generation \cite{chen2021unsupervised,zheng2021knowledge} so that the model can combine the most context-relevant knowledge to generate responses. The purpose of these works is to fuse context information and knowledge information through some appropriate methods. However, an essential and critical issue is that the knowledge is not provided in advance in reality. In other words, the above methods require pre-identified knowledge, but it is not feasible for a practical dialog system to acquire the knowledge needed to generate responses from the beginning. Hence, the retrieval of knowledge is a necessary step in practical applications. Although information retrieval has been proven effective in many fields, especially open-domain question answering (QA) \cite{karpukhin2020dense,chen2017reading,li2022mutually}, the response generation task is not similar to the QA task in which the question can be used as a natural query \cite{mao2021generation}. Therefore, naturally, for a multi-turn dialog session, we hope there is a method to dynamically spawn queries and utilize search engines to retrieve relevant knowledge.

On the other hand, numerous studies have generated meaningful responses using the provided knowledge and have shown satisfactory performance. However, most of them have not demonstrated their effectiveness under actual retrieval. 

\begin{figure}[!t]
	\includegraphics[width=1\columnwidth]{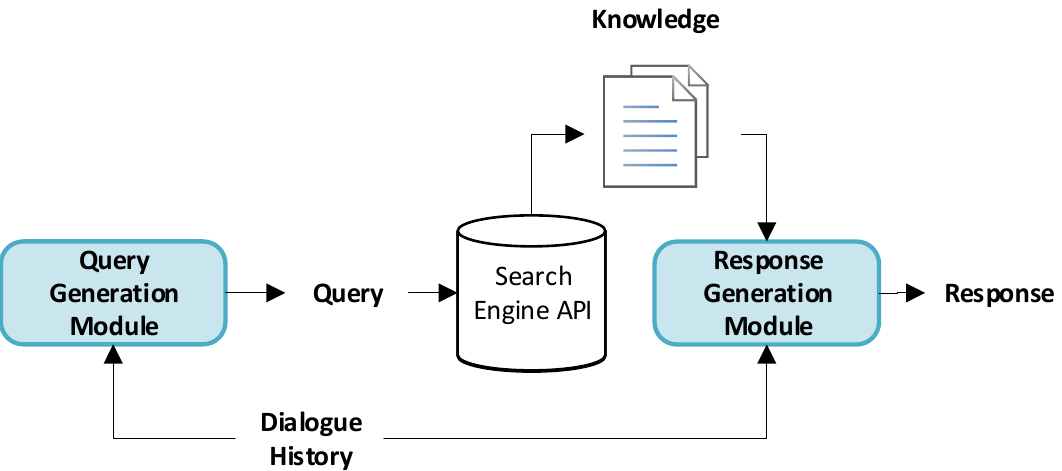}
	\caption{The high-level overview of our model.}
	\label{fig1}
\end{figure}

To alleviate the above problems and make the knowledge-driven dialog system more suitable for practical application scenarios, we propose a novel knowledge-driven dialog system named DRKQG. Specifically, as illustrated in Figure \ref{fig1}, the system can be divided into two simple but effective modules: the query generation module, which dynamically generates queries for retrieving knowledge as multi-turn dialog progresses, and the dialog generation module, which leads to informative and knowledgeable responses. For the query generation module, inspired by \cite{malhotra2022speaker}, we propose a time-aware mechanism that can guide the decoder to assign more attention to the more recent utterances.
Once the query is generated, a search engine can be called to retrieve relevant knowledge. Subsequently, for the response generation module, we combine the pointer network with the work of \cite{prabhumoye2021focused} so that the module can construct a context-driven representation of the knowledge and utilize knowledge-headed attention to attend over the representation in each decoder layer to generate or copy tokens.

In summary, our main contributions are as follows:
\begin{itemize}
	\item We propose the DRKQG model, which integrates query generation and response generation, enabling the dialog system to use search engine knowledge for open-domain dialog interaction, and further improving the practicality of knowledge-driven dialog systems.
	
	\item We employ some simple but effective methods to make our modules accomplish their respective tasks. For the query generation module, we develop a time-aware mechanism that can more reasonably assign attention weights to utterances. For the response generation module, we utilize a knowledge-headed attention to obtain representations of knowledge, and we copy tokens from context and knowledge to further improve the informativeness of generated response.
	
	\item Our experimental results in LIC2022 demonstrate that our model outperforms several baseline models. Moreover, human evaluation by Baidu's professional linguistics team shows that our model performs well in real-world.
\end{itemize}
\section{Related Work}
Due to the low performance of previous models, previous research on dialogue systems tended to focus on task-oriented dialogues \cite{JEONG201467,MATEJU2021327,https://doi.org/10.1111/coin.12544} and respond to users through reply choices \cite{liu2021filling,GU202112}. However, benefiting from the development of deep learning techniques, especially the prosperity of pre-trained language models in recent years such as Bert \cite{devlin2019bert} and Bart  \cite{lewis2020bart}, remarkable achievements have been made in open-domain response generation task \cite{ide2021multi, wang2021fcm,LIU2021106}. The task can generally be regarded as a natural language generation (NLG) task that produces satisfactory responses by giving the dialog history. However, these dialog systems can always generate dull and meaningless responses due to the inherent flaws of NLG tasks \cite{xu2021adaptive}. 

To tackle the above problems, dialog systems with multi information source, such as \cite{tian2019learning,wang2020improving,2021Towards, zhao2023multi} have recently received widespread attention. Their experiments proved that extra information could enhance the performance of response generation. Among them, using knowledge as an external information source is the most common \cite{GU202113,SHU2023}, because previous dialogue systems are lacking in factual accuracy and knowledge richness. Several works focus on fusing knowledge information with dialog context. For example, \cite{prabhumoye2021focused} used additional multi-head attention to obtain the representation of knowledge information, and \cite{cao2020pretrained} proposed an attention fusion mechanism to fuse document information and context information. Other works integrated knowledge selection with response generation \cite{chen2021unsupervised,zheng2021knowledge,lianlearning}. Specifically, these methods filter knowledge irrelevant to the context and then generate responses based on the selected knowledge. Moreover, \cite{wu2019proactive,liu2020towards,bai2021learning} believe that the goal of a dialog session can provide implicit guidance for knowledge-driven dialog systems. The above works all assume that knowledge will be provided in advance; however, this assumption is not valid for a natural dialog system. Therefore, our work focuses on allowing the dialog system to automatically acquire relevant knowledge through some methods, such as retrieval through search engines.

Some other methods also guide our work. For example, we use a pointer network \cite{vinyals2015pointer} to copy tokens from context and knowledge to further improve the knowledgeable of the generated response. Moreover, inspired by \cite{malhotra2022speaker}, we propose a time-aware mechanism to capture context information so that the more recent utterances can receive more attention.
\begin{figure*}[htbp]
	\centering
	\includegraphics[width=2\columnwidth]{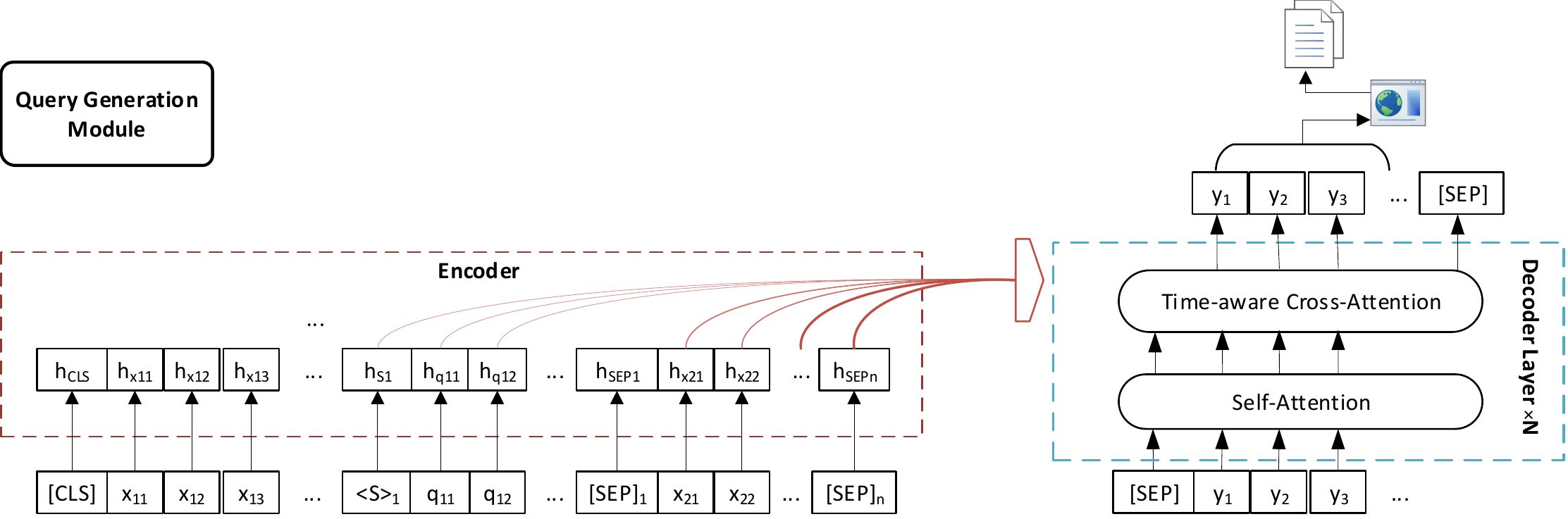}
	\caption{The architecture of our query generation module. Here, the time-aware attention mechanism will pay more attention to the later dialogs in the dialog history, and we use lines with different shades of color to represent the attention weights.}
	\label{fig2}
\end{figure*}
\section{Model Description}
\subsection{Task Formalization}
Suppose we have a dataset $D = \{ ({U_i},{Y_i},{Q_i},{K_i})\} _i^N$ where ${U_i} = (u_i^1,u_i^2,...,u_i^{l_u})$ represents the dialog history, ${Q_i} = (q_i^1,q_i^2,...,q_i^{l_q})$ is the query produced by dialog history ${U_i}$, and ${K_i} = (k_i^1,k_i^2,...,k_i^{l_k})$ is the knowledge relevant to the context, which is provided by the dataset during training but retrieved by Q in reality. The response ${Y_i} = (y_i^1,y_i^2,...,y_i^{l_y})$ should be generated by the context ${U_i}$ and the relative knowledge ${K_i}$. Here, $N$ represents the size of the dataset, and $l_u$, $l_q$, $l_k$ and $l_y$ denote the numbers of tokens in the dialog history, query, knowledge and response, respectively. The final goal of our proposed system is to generate a response on the ground of the context and knowledge. In the training stage, we train our query generation module and response generation module separately.
\subsection{Model Overview}
The DRKQG consists of two modules, a query generation module and a response generation module, as illustrated in Figure \ref{fig1}.
The query generation module can generate a suitable query, and then a search engine can be called to retrieve relevant knowledge. Here, the search engine API is provided by LIC2022. Subsequently, the response generation module can generate an informative response based on the context and the retrieved knowledge. The main framework of each module is a transformer-based \cite{vaswani2017attention} encoder-decoder (CPT \cite{shao2021cpt}), and each module has its additional mechanisms to make them better adapted to their respective tasks. Below, we first introduce the query generation module we proposed (Section \ref{sec3.3}), especially the time-aware mechanism applied to it. Then, we introduce the response generation module (Section \ref{sec3.4}), which utilizes additional knowledge-headed attention and copy mechanism to produce a response. Finally, we detail the training strategy (Section \ref{sec3.5}).
\subsection{Query Generation Module}\label{sec3.3}
In this section, we will introduce our proposed query generation model, which combines a unique time-aware mechanism so that the later utterances in the dialog history are assigned greater attention weights, as shown in Figure \ref{fig2}.
\subsubsection{Context Representation}
We concatenate the utterances in dialog history and their corresponding generated queries, denoted as $X = (utteranc{e^1};quer{y^1};...;utteranc{e^n})$ and we observe that for the same dialog session, the queries that have been generated should have a certain prompting effect on the query to be generated. Then, we take $X$ as the input of the encoder and obtain the representation $H_x$ of the context via the CPT encoder:
\begin{equation}
	{H_x} = Encoder_{q}(X);H_x \in \mathbb{R}^{l \times d}\label{e1}\nonumber
\end{equation}
where $l$ is the length of the encoder input. Likewise, we obtain the representation $H_q$ of the query $Q$ by:
\begin{equation}
	{H_q} = Encoder_{q}(Q);H_q \in \mathbb{R}^{l_q \times d}\label{e2}\nonumber
\end{equation}
where $l_q$ is the length of the decoder input. The impact of corresponding generated queries will be discussed in Section \ref{sec5.1} and \ref{a}.
\subsubsection{Time-aware Attention}
Intuitively, the topic of the conversation will change as the conversation progresses, and we humans generally only focus on the topic just discussed. Therefore, the purpose of generating a query is usually to retrieve knowledge and use it to reply to what the user has just said. Inspired by \cite{malhotra2022speaker}, we apply a \emph{time-aware attention mechanism} that can learn the importance of utterances according to the order in which they appear in the dialog history:
\begin{equation}
	\begin{gathered}
		{Attention_{T}} = Softmax(t\frac{{Q{K^{T}}}}{{\sqrt d }})V;t \in {\mathbb{R}^l},{t^i} = {N_u}(i)\hfill \\
		{N_u}(i) = j;{X^i} \in [utteranc{e^j};quer{y^j}]\nonumber
	\end{gathered}
\end{equation}
Here, $Q$, $K$, and $V$ represent the $query$, $key$ and $value$ in an $Attention$ function respectively, and the $t$ continues to increase with the number of dialog turns. In this way, the weight of the dot product of the query and key increases with time, and we can obtain the time-aware representations of the context.

The original time-aware mechanism in \cite{malhotra2022speaker} uses the memory network to store the encoder hidden state for each utterance, and then exploits the nearby hidden states maintained in the memory to obtain the representation at every point in the dialog. However, our \emph{time-aware mechanism} does not obtain the representation on the encoder side for each utterance based on their nearby context, but enables the decoder to allocate attention to the dialog history more reasonably without additional parameters.
\subsubsection{Decoder}
Now, each decoder layer consists of the following functions:
\begin{equation}
	\begin{gathered}
		H = LayerNorm(SelfAttention({H_q},{H_q},{H_q})) \hfill \\
		H = LayerNorm(Attentio{n_T}(H,H_x,H_x)) \hfill \\
		H = LayerNorm(FFN(H)) \hfill \nonumber
	\end{gathered}
\end{equation}
We replace $CrossAttention$ in the Transformer Decoder-block with our proposed \emph{time-aware attention mechanism}. Compared with general $CrossAttention$, this mechanism of making the decoder pay more attention to the more recent conversations allows the model to generate a reasonable query.
\subsection{Response Generation Module}\label{sec3.4}
\begin{figure*}[htbp]
	\centering
	\includegraphics[width=2\columnwidth]{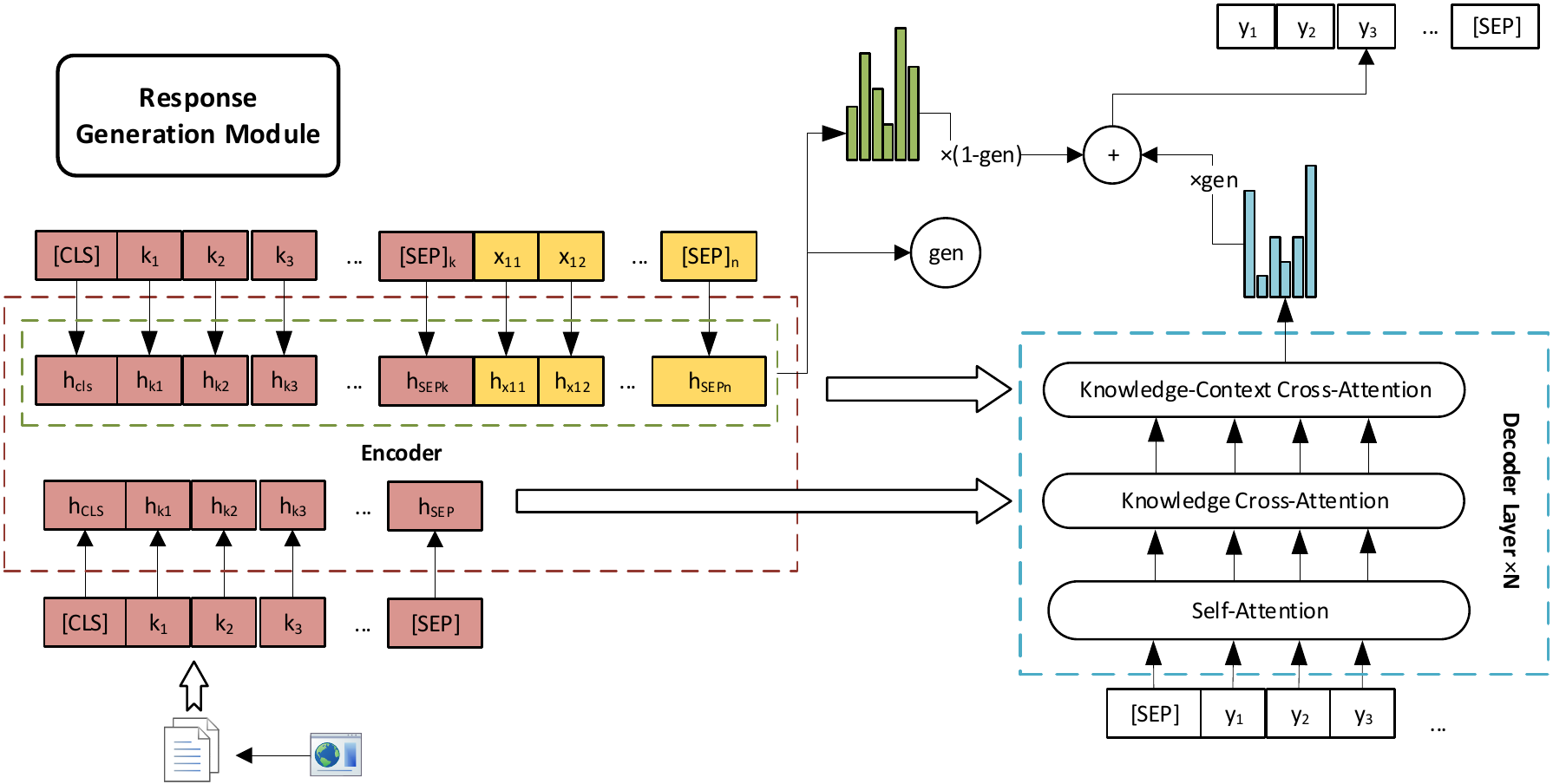}
	\caption{The architecture of our response generation module, which utilizes a knowledge-headed attention to attend over the representation in each decoder layer, and the module also uses a pointer network to copy words from context and knowledge.}
	\label{fig3}
\end{figure*}
Once the query is generated, we can obtain knowledge through the search engine API. Next, we introduce our response generation module in this section, which can use the retrieved knowledge and context to generate a knowledgeable response, as shown in Figure \ref{fig3}.
\subsubsection{Context Representation}
Inspired by the work of \cite{prabhumoye2021focused}, we concatenate the retrieved knowledge and the context so that we can get our knowledge-context representation $H_{kc}$ by:
\begin{equation}
	{H_{kc}} = Encoder_{res}([K;X]);H_{kc} \in \mathbb{R}^{{(l_k+l)} \times d}\label{e5}\nonumber
\end{equation}
and we then utilize the same CPT encoder to obtain the representation of the knowledge alone:
\begin{equation}
	{H_{k}} = Encoder_{res}(K);H_{k} \in \mathbb{R}^{{l_k} \times d}\label{e6}\nonumber
\end{equation}
\subsubsection{Knowledge-Headed Attention}
Usually, a transformer decoder block contains a \emph{SelfAttention} layer and a \emph{CrossAttention} layer. The \emph{SelfAttention} layer allows each token in the decoder to assign attention to its position and the position before it. In contrast, the \emph{CrossAttention} layer enables the source sequence to be aware of the encoder output, allowing their representations to be fused. In this module, we have two encoder outputs, representing context-driven information of the knowledge and the knowledge information, respectively. Therefore, we add an additional \emph{CrossAttention} layer (called $CrossAttentio{n_{{\text{kc}}}}$), which attends over the knowledge and context. At the same time, the original \emph{CrossAttention} layer (called $CrossAttentio{n_{{\text{k}}}}$) can focus only on knowledge information:
\begin{equation}
	CrossAttentio{n_{{\text{kc}}}} = Attention(H,{H_{kc}},{H_{kc}})\nonumber
\end{equation}
\begin{equation}
	CrossAttentio{n_{{\text{k}}}} = Attention(H,{H_{k}},{H_{k}})\nonumber
\end{equation}
\subsubsection{Decoder}
Now, each decoder layer in this module consists of the following functions:
\begin{equation}
	\begin{gathered}
		H = LayerNorm(SelfAttention({H_y},{H_y},{H_y})) \hfill \\
		H = LayerNorm(CrossAttentio{n_{{\text{k}}}}({H},{H_k},{H_k})) \hfill \\
		H = LayerNorm(CrossAttentio{n_{{\text{kc}}}}({H},{H_{kc}},{H_{kc}})) \hfill \\
		H = LayerNorm(FFN(H)) \hfill \nonumber
	\end{gathered}
\end{equation}
In this way, the module can fully extract knowledge and contextual features. We call the method KoHAC, and the difference between KoHAC and the model DoHA from \cite{prabhumoye2021focused} will be discussed in Section \ref{sec5.4}. Finally, the probability distributions of token generation $P_v$ can be obtained by:
\begin{equation}
	{P_v} = Softmax(W_{head}^{T}H + b)\nonumber
\end{equation}
where $W_{head}$ and $b$ are learnable parameters.
\subsubsection{Pointer Network}
To further improve the consistency and accuracy of knowledge in the generated responses, we also incorporate a pointer network into this module. Unlike the standard pointer network, we did not consider the oov problem and used a simple way to combine the copy mechanism with the language model. Specifically, the decoder contains N decoding layers, and we take the mean of the attention distributions of these decoding layers over the knowledge-context representation as our pointer. Thus, each token generation can be determined with a soft gate that can control the probability of generation or copy:
\begin{equation}
	gen = \sigma (W_s^TH)\nonumber
\end{equation}
where $W_s^{T}$ is a learnable parameter. Finally, the probability of the vocabulary can be defined as:
\begin{equation}
	P = gen \times {P_v} + ((1 - gen) \times {P_{copy}})\nonumber
\end{equation}
where $P_{copy}$ is the mean of the attention distributions of all decoding layers over the knowledge-context representation.
\subsection{Training}\label{sec3.5}
We train the two modules separately. The query generation module uses the dialog history and previously used queries to generate an appropriate new query, while the response generation module generates responses based on the currently retrieved knowledge and context. During the training phase, the dialog history, previously used queries, and knowledge are provided by the dataset:
\begin{equation}
	\begin{gathered}
		{L_{query}}(\theta _1) =  - \frac{1}{{\left| Q \right|}}\sum\limits_{t = 1}^{\left| Q \right|} {\log (P({q^t}} |{q^{1:t - 1}},U,{Q_{past}})) \hfill \\
		{L_{res}}({\theta _2}) =  - \frac{1}{{\left| Y \right|}}\sum\limits_{t = 1}^{\left| Y \right|} {\log (P({y^t}} |{y^{1:t - 1}},U,K)) \hfill \nonumber
	\end{gathered}
\end{equation}
where ${Q_{past}}$ represents the queries that have been produced in the same dialog session.

\section{Experiments}
\subsection{Dataset}
We used the \textbf{Dusinc} \cite{zhou2022sinc} dataset provided in LIC2022\footnote{aistudio.baidu.com/aistudio/competition/detail/158/0}, Language and Intelligence Technology Competition which Baidu hosts. As far as we know, there is no other Chinese dataset suitable for query and response generation tasks in knowledge-driven dialog systems.
\subsubsection{Dusinc}
The DuSinc dataset is an open-domain Chinese dialog dataset that contains a wide range of dialog topics from real human conversations. During the data collection process, the dialog participants are required to play the roles of the USER and the BOT. The BOT can query the search engine in real-time during the chat and conduct in-depth dialog and interaction with the USER. If the BOT does not use extra knowledge to generate the response, the query and knowledge provided in this round of dialog will be empty. The dataset consists of 2200 dialog sessions with 11466 turns, 10413/1053 used for train/valid, and each turn of BOT includes the corresponding query and knowledge. Therefore, we can train our modules separately. What is more, there are another 350/750 samples for testing query/response generation.
For the cases where the query and knowledge are empty, we discuss them in \ref{b}.
\subsection{Comparison Models}
We implement our model on the Dusinc dataset. However, most of the existing models cannot handle both the query and response generation tasks, and they did not perform any experiments on Dusinc. Consequently, we implement these comparison models for automatic evaluation metrics and compare each module of our system separately with the corresponding models.
\subsubsection{Query Generation}
Few studies focus on query generation in knowledge-driven dialogs. Thus, We compare our query generation module with: \textbf{Transformer Encoder\footnote{github.com/PaddlePaddle/Knover/tree/dygraph/projects/-lic2022}:} This model is a baseline model provided by LIC2022, which consists of 12 layers of transformer encoder and is pre-trained on a large Chinese dialog dataset. \textbf{CPT-large:}  CPT is a pre-trained language model that we used in our module, and it has achieved outstanding results on several Chinese datasets.
\subsubsection{Response Generation}
We compare our Response generation module with: \textbf{Transformer Encoder:} This model is a baseline model provided by LIC2022. \textbf{CPT-large:} The CPT is a pre-trained language model that we used in our module. \textbf{CPT-KIC:} The KIC model is proposed by \cite{lin2020generating}, and it has achieved impressive results on several knowledge-based dialog datasets. We implement the recurrent knowledge interaction and knowledge copy method in the KIC model with the CPT. \textbf{CPT-sw:} The original model GPT2-sw was proposed by \cite{cao2020pretrained}, which can generate responses with multiple input sources. Here, we make minor changes to the original method for information fusion of multiple input sources; thus, the method can be applied in an encoder-decoder model. \textbf{DoHA:} a model that can build knowledge-context representation and implement specific attention to the knowledge-context \cite{prabhumoye2021focused}, and the KoHAC we proposed is improved on the basis of it.
\begin{table*}[!t]
	\centering
	\begin{tabular*}{\hsize}{@{}@{\extracolsep{\fill}}ccccc@{}}
		\toprule
		\textbf{Model}      & \textbf{Q\_F1} & \textbf{Q\_BLEU-1} & \textbf{Q\_BLEU-2}  & \textbf{Parameters}\\ \midrule
		\textbf{Ours(CPT-Time)}                & 0.471       & 0.416           & 0.370  & 407M         \\ 
		Transformer Encoder & 0.163       & 0.108           & 0.106          \\
		CPT-Large*          & 0.424       & 0.351           & 0.317      & 407M\\ \bottomrule
	\end{tabular*}
	\caption{The experimental results for query generation. * means that the comparison model is implemented by ourselves.}
	\label{tab1}
\end{table*}
\begin{table*}[!t]
	\centering
	\begin{tabular*}{\hsize}{@{}@{\extracolsep{\fill}}cccccc@{}}
		\toprule
		\textbf{Model}      & \textbf{D\_F1} & \textbf{D\_BLEU-1} & \textbf{D\_BLEU-2} & \textbf{D\_DISTINCT-1} & \textbf{D\_DISTINCT-2}\\ \midrule
		\textbf{Ours(KoHAC)}       & 0.333          & 0.308    & 0.216     &  0.117 & 0.588          \\
		Transformer Encoder & 0.2            & 0.137  & 0.088          & 0.148 & 0.535          \\
		CPT-Large*         & 0.295          & 0.246  & 0.171          & 0.173 & 0.660          \\
		CPT-KIC*         & 0.283          & 0.223  & 0.143          & 0.075  &0.392          \\
		CPT-sw*              & 0.303          & 0.237   &0.163          & 0.164  &0.641          \\
		CPT-DoHA*            & 0.314          & 0.255  &0.184          & 0.168   &0.671           \\ \bottomrule
	\end{tabular*}
	\caption{The experimental results for response generation. * means that the comparison model is implemented by ourselves.}
	\label{tab2}
\end{table*}
\subsection{Metrics}
For performance evaluation, we utilize automatic evaluation and human evaluation.
\subsubsection{Automatic Evaluation}
Following the setting of LIC2022, we use F1, BLEU1/2 \cite{sutskever195sequence}, and DISTINCT1/2 \cite{li2015diversity} to evaluate our query generation module and response generation module. Specifically, Q\_F1 indicates that the word granularity matching score between the predicted query and the golden query; Q\_BLEU represents the BLEU1/2 value of the word granularity between the predicted query and the golden query, estimating the fluency of the generated query; D\_F1 can measure the word granularity matching score between the predicted response and the golden response; D\_BLEU represents the BLEU1/2 value of the word granularity between the predicted response and the golden response; D\_DISTINCT1/2 is an automatic evaluation metric of the diversity of response content.
\subsubsection{Human Evaluation}
Human evaluations are provided by Baidu's professional linguistics team, which tests our system in real-world scenarios. Specifically, the team will use pre-designed dialog topics to interact with our system. For a dialog session, our model needs to dynamically generate a query and retrieve knowledge through a search engine API which LIC2022 provides, and then produce a response based on the retrieved knowledge in each round of the dialog session. There will be 30 dialog sessions used to test our system, and each dialog session will not be less than 7 turns of dialog. Ultimately, the following four metrics are used to evaluate the quality of our generated responses.
\begin{itemize}
	\item Knowledgeable (0$\sim$1): Evaluate the knowledgeable of generated responses.
	\item Factually Correct (0$\sim$1): Evaluate the factual accuracy of the knowledge used.
	\item Consistent (0$\sim$1): Evaluate the appropriateness of the responses to the context.
	\item Engaging (0$\sim$1): whether the responses are attractive to the chatter.
\end{itemize}
\subsection{Implementation}
We implement our model with Pytorch. CPT-large is used as the backbone of our model, which consists of 20 layers of transformer encoder and 4 layers of transformer decoder. The hidden size is 1024, and the vocabulary size is 21128. We use the Adam optimizer \cite{kingma2014adam} and train each module on two GPUs (RTX TITAN); the batch size is 12, and the learning rate is $4e^{-5}$. What is more, we apply a scheduler that warms up the training with 200 steps. All the above settings are the same for both modules.

In particular, the knowledge and the dialog are encoded by a shared CPT-Encoder in the response generation module, and we implement the pointer network inspired by \cite{vinyals2015pointer}. The sequence length for knowledge and context are both 128, the maximum sequence length of output is 128 and we use 2 epochs to train the response generation module. However, for the query generation module, and the maximum sequence length of output is 16 and we train it with 3 epochs.
\subsection{Results} 
\subsubsection{Automatic Evaluation}
The automatic evaluation\footnote{The results of the automatic evaluation metrics in this paper are slightly inferior to those presented in LIC2022. Because we added a post-processing step that is not used in this paper and in the human evaluation stage to correct the results in the automatic evaluation stage of LIC2022.} results of the \textbf{query generation module} on the Dusinc dataset are shown in Table \ref{tab1}. Our query generation module outperforms the baseline model by a large margin over all evaluation metrics, and it also achieves significant improvement  without additional parameters compared to CPT-Large, which powerfully demonstrates the effectiveness of the \emph{time-aware mechanism} we proposed. Specifically, our query generation module has 11.1\%, 18.5\% and 16.7\% improvement over CPT-Large on F1, BLEU-1, and BLEU-2, which indicates that our \emph{time-aware mechanism} can effectively capture clues from more recent conversations and generate more accurate and fluent queries than general attention mechanisms that automatically assign weights to the entire dialog history.
\begin{figure*}[htbp]
	\centering
	\includegraphics[width=2\columnwidth]{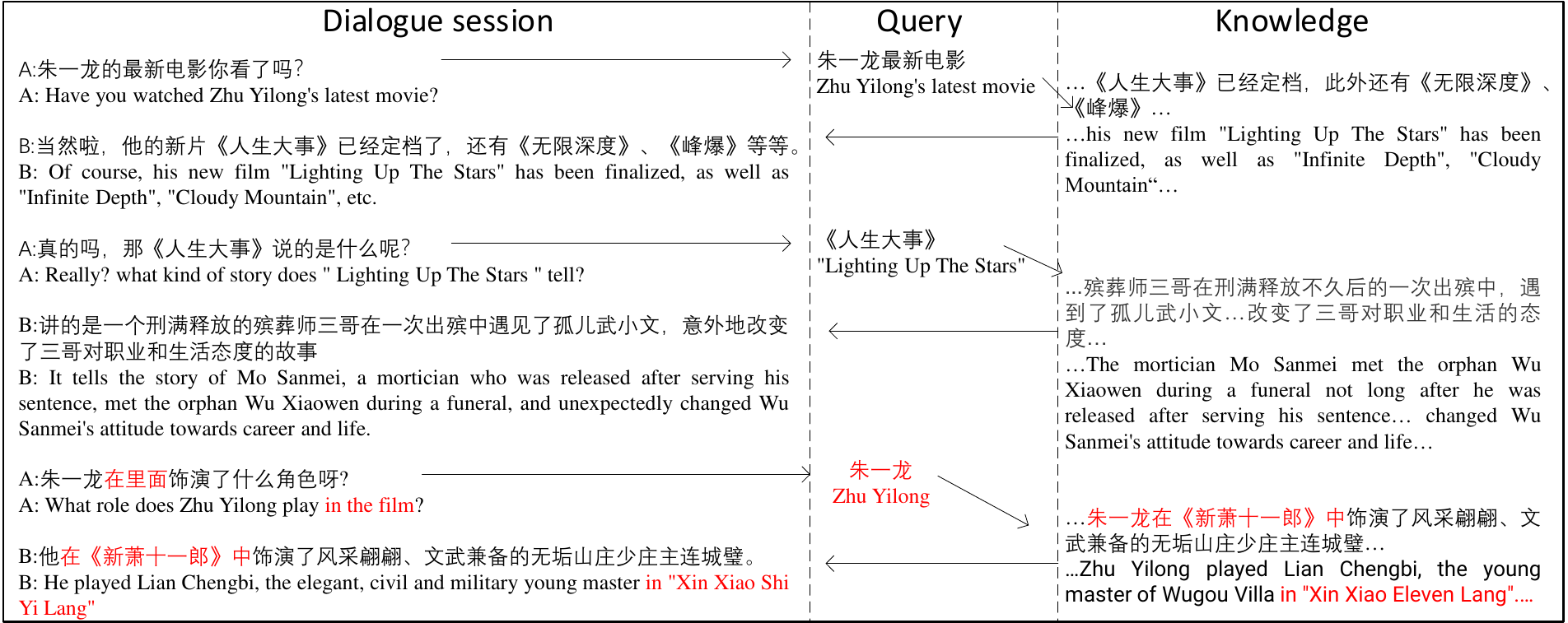}
	\caption{An example of multi-turn communication between our dialog system and a tester. There are a total of 6 turns of dialog in this session. After the user enters a sentence, a query will be generated based on the context and used to retrieve relevant knowledge, and then a response will be produced derived from the retrieved knowledge and context.}
	\label{fig4}
\end{figure*}

The automatic evaluation results of the \textbf{response generation module} on the Dusinc dataset are shown in Table \ref{tab2}. Our response generation module outperforms other comparison models over most automatic metrics. Specifically, our response generation module has 6.0\%, 20.7\% and 17.0\% improvement over CPT-DoHA on F1, BLEU-1, and BLEU-2, respectively, but drops by 30.3\% and 12.3\% on DISTINCT-1 and DISTINCT-2. The above results show that combining the pointer network with the pre-trained language model can indeed make the knowledge-driven dialog system generate more knowledge-coherent responses because the pointer network will directly copy words from the knowledge to the output. However, the copy mechanism also inevitably limits the diversity of responses generated. What surprises us in this result is that the metrics of CPT-KIC have decreased compared with CPT-Large. We conjecture that this is because the recurrent knowledge injection method proposed in KIC needs to select appropriate knowledge in each decoding step, and it cannot handle the samples well where the knowledge provided by the Dusinc dataset is empty.
\subsubsection{Human Evaluation}
In human evaluation, our system needs to strictly perform the steps of query generation, knowledge retrieval, and response generation. The evaluator will evaluate the quality of the final generated response in the field of Knowledgeable, Factually Correct, Consistent, and Engaging. We compare our results with those of the other top ten contestants in LIC2022, as shown in Table \ref{tab3}. The responses generated by our system have achieved outstanding results on Knowledgeable and Factually Correct, indicating that our query generation module can generate appropriate queries to retrieve context-relevant knowledge, and our response generation module can effectively utilize the retrieved knowledge to generate informative responses. However, our generated responses are less consistent and engaging than some contestants. The results on Consistent and Engaging are not surprising to us, as our primary work is to make knowledge-driven dialog systems leverage relevant knowledge in real-world settings. We have not taken any optimization measures for the diversity and consistency of response generation. In the following subsection, we further illustrate the strengths and weaknesses of our model with a case.
\begin{table}[!t]
	\centering
	\resizebox{\linewidth}{!}{
		\begin{tabular}{@{}ccccc@{}}
			\toprule
			\textbf{Team} & \textbf{Knowledgeable} & \textbf{Factually correct} & \textbf{Consistent} & \textbf{Engaging} \\ \midrule
			\textbf{Ours} & \textbf{0.627}         & \textbf{0.607}             & \textbf{0.639}      & \textbf{0.641}    \\
			Team-A        & 0.526                  & 0.481                      & 0.855               & 0.869             \\
			Team-B        & 0.712                  & 0.674                      & 0.786               & 0.81              \\
			Team-C        & 0.568                  & 0.526                      & 0.755               & 0.804             \\
			Team-D        & 0.486                  & 0.423                      & 0.84                & 0.854             \\
			Team-E        & 0.568                  & 0.518                      & 0.7                 & 0.668             \\
			Team-F        & 0.621                  & 0.531                      & 0.649               & 0.673             \\
			Team-G        & 0.514                  & 0.454                      & 0.645               & 0.671             \\
			Team-H        & 0.495                  & 0.435                      & 0.617               & 0.581             \\
			Team-I        & 0.284                  & 0.235                      & 0.564               & 0.543             \\ \bottomrule
		\end{tabular}
	}
	\caption{The result of human evaluation in no particular order. Compared with other top contestants, our final generated responses are outstanding in terms of Knowledgeable and Factually correct.}
	\label{tab3}
\end{table}

\section{Analysis and discussions}
\subsection{Ablation Study}\label{sec5.1}
In order to verify the effectiveness of our proposed methods, we perform ablation study on the query generation module and the response generation module, respectively. We conduct a series of ablation experiments by removing different components of the modules, and the results are shown in Table \ref{tab4} and Table \ref{tab5}.
\begin{table*}[!t]
	\centering
	\begin{tabular*}{\hsize}{@{}@{\extracolsep{\fill}}cccc@{}}
		\toprule
		\textbf{Model}      & \textbf{Q\_F1} & \textbf{Q\_BLEU-1} & \textbf{Q\_BLEU-2} \\ \midrule
		\textbf{Ours(CPT-Time)}                & 0.471       & 0.416           & 0.370          \\ 
		w/o Time-aware          & 0.424       & 0.351           & 0.317 \\
		w/o Generated queries  & 0.464       & 0.379           & 0.348          \\
		w Retrieved knowledge          & 0.448       & 0.324           & 0.298       \\ \bottomrule
	\end{tabular*}
	\caption{The ablation study for our CPT-time.}
	\label{tab4}
\end{table*}
\begin{table*}[!t]
	\centering
	\begin{tabular*}{\hsize}{@{}@{\extracolsep{\fill}}cccccc@{}}
		\toprule
		\textbf{Model}      & \textbf{D\_F1} & \multicolumn{2}{c}{\textbf{D\_BLEU-1/2}} & \multicolumn{2}{c}{\textbf{D\_DISTINCT-1/2}} \\ \midrule
		\textbf{KoHAC}       & 0.333          & \multicolumn{2}{c}{0.308/0.216}          & \multicolumn{2}{c}{0.117/0.588}          \\
		w/o Knowledge-headed         & 0.317          & \multicolumn{2}{c}{0.274/0.191}          & \multicolumn{2}{c}{0.115/0.563}          \\
		w/o copy            & 0.317          & \multicolumn{2}{c}{0.266/0.188}          & \multicolumn{2}{c}{0.159/0.673}           \\ 
		w/o both        & 0.295          & \multicolumn{2}{c}{0.246/0.171}          & \multicolumn{2}{c}{0.173/0.660}          \\ \bottomrule
	\end{tabular*}
	\caption{The ablation study for our KoHAC. w/o Knowledge-headed means we remove $CrossAttention_{kc}$ and we concat dialogue history and knowledge as our input sequence. w/o both means we remove knowledge-headed attention and copy mechanism.}
	\label{tab5}
\end{table*}
For the query generation module, we first remove the time-aware, so that our query generation module degenerates into a standard CPT-Large model. It can be clearly seen that the performance of all the metrics has dropped significantly, which proves that the time-aware plays an crucial role in the query generation module. What is more, it also demonstrates our point that the interlocutor usually pays more attention to the topic just discussed.

Second, we study the impact of context content on query generation module performance. Specifically, we explore the effect of generated queries and retrieved knowledge on the current query generation. The original context we use is the splicing of utterances in the dialogue history and the corresponding generated queries, denoted as $X = (utteranc{e^1};quer{y^1};...;utteranc{e^n})$. We remove the generated queries so that $X^{-query} = (utteranc{e^1};...;utteranc{e^n})$ and we get $X^{+knowledge} = (utteranc{e^1};knowledg{e^1};...;utteranc{e^n})$ by concatenating the utterances and their corresponding retrieved knowledge. We observe that for the same dialog session, the queries that have been generated should have a certain prompting effect on the query to be generated. However, it hurts performance when the context is incorporated into retrieved knowledge, which we believe is due to the fact that the retrieved knowledge is too long and introduces a lot of noise.

For the response generation module, we first perform ablation study for the knowledge-headed attention and copy network, and we found that both the knowledge-headed attention and the copy network improved the performance of the module on the F1 and BLEU. However, the performance of the model on DISTINCT has improved when we remove the copy network, because the copy mechanism can only select tokens from knowledge and dialogue history, it will inevitably drop the diversity of response. Then, poor F1 and BLEU scores are obtained if we remove both of the knowledge-headed attention and the copy mechanism.

\subsection{Time-aware vs. position embedding}
\begin{table*}[!t]
	\centering
	\begin{tabular*}{\hsize}{@{}@{\extracolsep{\fill}}cccc@{}}
		\toprule
		\textbf{Model}      & \textbf{Q\_F1} & \textbf{Q\_BLEU-1} & \textbf{Q\_BLEU-2} \\ \midrule
		\textbf{Ours(CPT-Time)}                & 0.471       & 0.416           & 0.370          \\ 
		w/o Time-aware          & 0.424       & 0.351           & 0.317 \\
		w/o position  & 0.38       & 0.301           & 0.258          \\
		w/o both          & 0.286       & 0.233           & 0.185       \\ \bottomrule
	\end{tabular*}
	\caption{The experimental results for comparing time-aware and position embedding.  w/o both means we remove time-aware attention and position embedding.}
	\label{tab6}
\end{table*}

Position embedding can also learn time information along with training, however, the position embedding and the time-aware are still fundamentally different. Specifically, the position embedding captures time information at the token-level while the time-aware attention forces the model to assign attention weights according to the order of utterance and capture time information at the utterance-level. We verify the impact of the position embedding and the time-aware on query generation module, as shown in Table \ref{tab6}.

It can be seen that regardless of removing the position embedding or removing the time-aware, the performance on all the metrics has dropped significantly. what is more, when we remove the position embedding and the time-aware attention at the same time, the performance of the query generation module is even worse. Therefore, we can believe that time information is crucial in the query generation module, and the time-aware and position embedding can capture time information from different level.
\subsection{Time-aware vs. Last utterance}
Intuitively, the topic of the conversation will change as the conversation progresses, and the generated query is often related to the recently discussed topic. Therefore, we propose the time-aware attention mechanism. However, it is worth exploring whether the conversations that occurred earlier have any impact on the current query to be generated. Specifically, we only use the last utterance of the dialogue history as the input of the query generation module, and the experimental results are shown in the Table \ref{tab7}.
\begin{table}[!t]
	\centering
	\resizebox{\linewidth}{!}{
		\begin{tabular}{@{}cccc@{}}
			\toprule
			\textbf{Model}      & \textbf{Q\_F1} & \textbf{Q\_BLEU-1} & \textbf{Q\_BLEU-2} \\ \midrule
			\textbf{Ours(CPT-Time)}                & 0.471       & 0.416           & 0.370          \\ 
			Last utterance          & 0.458       & 0.391           & 0.348 \\ \bottomrule
		\end{tabular}
	}
	\caption{The experimental results for Time-aware vs. Last utterance}
	\label{tab7}
\end{table}

Obviously, using only the last sentence as the information source will drop the experimental performance to a certain extent. This shows that the query generation module needs to reasonably allocate attention to the utterances in the dialogue history and extract effective information from the entire conversation, instead of relying only on the last utterance, even if it is indeed significant.
\subsection{KoHAC vs. DoHA}\label{sec5.4}
DoHA has indeed made remarkable achievements, but it is not perfect for the knowledge dialogue system based on Internet retrieval. First, the knowledge retrieved from the Internet is verbose, and its length is often several times that of the dialogue history. Second, knowledge-driven dialogue systems focus more on the accuracy of the generated knowledgeable responses. However, the encoding mode of DoHA results in it paying more attention to dialogue history information than knowledge information. The DoHA method will be presented below.
\paragraph{Context representation of DoHA}
First DoHA concatenate the dialogye and the document so that it can get the context-document representation $H_{cd}$ by:
\begin{equation}
	{H_{cd}} = Encoder([X;D])\nonumber
\end{equation}
and it then utilize the same encoder to obtain the representation of the dialogue history alone:
\begin{equation}
	{H_{c}} = Encoder(X)\nonumber
\end{equation}
\paragraph{Document-Headed Attention of DoHA}
It has two \emph{CrossAttention} layers, one called $CrossAttentio{n_{{\text{cd}}}}$, which attends over the document and context, and the other called $CrossAttentio{n_{{\text{c}}}}$ (the original \emph{CrossAttention} layer), which can focus only on context information:
\begin{equation}
	CrossAttentio{n_{{\text{cd}}}} = Attention(H,{H_{cd}},{H_{cd}})\nonumber
\end{equation}
\begin{equation}
	CrossAttentio{n_{{\text{c}}}} = Attention(H,{H_{c}},{H_{c}})\nonumber
\end{equation}
\paragraph{Decoder of DoHA}
Now, each decoder layer in DoHA consists of the following functions:
\begin{equation}
	\begin{gathered}
		H = LayerNorm(SelfAttention({H_y},{H_y},{H_y})) \hfill \\
		H = LayerNorm(CrossAttentio{n_{{\text{c}}}}({H},{H_c},{H_c})) \hfill \\
		H = LayerNorm(CrossAttentio{n_{{\text{cd}}}}({H},{H_{cd}},{H_{cd}})) \hfill \\
		H = LayerNorm(FFN(H)) \hfill \nonumber
	\end{gathered}
\end{equation}
In this way, the DoHA can fully extract document and contextual features. 

KoHAC made some modifications on the basis of DoHA, making the model pay more attention to knowledge information, as described in Section \ref{sec3.4}. The experimental results are shown in the Table \ref{tab8}.
\begin{table}[!t]
	\centering
	\resizebox{\linewidth}{!}{
		\begin{tabular}{@{}cccccc@{}}
			\toprule
			\textbf{Model}      & \textbf{D\_F1} & \multicolumn{2}{c}{\textbf{D\_BLEU-1/2}} & \multicolumn{2}{c}{\textbf{D\_DISTINCT-1/2}} \\ \midrule
			\textbf{Ours(KoHAC w/o copy)}                & 0.317           & \multicolumn{2}{c}{0.266/0.188}          & \multicolumn{2}{c}{0.159/0.673} \\
			CPT-DoHA*             & 0.314           & \multicolumn{2}{c}{0.255/0.184}          & \multicolumn{2}{c}{0.168/0.671}          \\
			remove extra            & 0.295         & \multicolumn{2}{c}{0.246/0.171}          & \multicolumn{2}{c}{0.173/0.660}           \\ \bottomrule
		\end{tabular}
	}
	\caption{The experimental results for KoHAC vs. DoHA. Remove extra means we remove the extra headed attention, and we concat dialogue history and knowledge as our input sequence.}
	\label{tab8}
\end{table}

\subsection{Case Study}
We present an example of multi-turn communication between our dialog system and a tester, as illustrated in Figure \ref{fig4}. Speaker A is our tester, and speaker B represents our dialog system. Speaker A first confirms a dialog topic and continuously constructs new inputs to guide the conversation. The dialog system dynamically produces queries and generates responses derived from the retrieved knowledge and the context. In this example, the tester and our dialog system discussed an actor and the movie he was in. Our model generated accurate queries and satisfactory responses in the previous four turns, demonstrating that both proposed modules can complete their respective tasks well, and our system DRKQG performs well in the actual environment. However, our query generation module performed poorly when generating the query for the 5th turn of dialog. It generated only the actor's name but not the title of the corresponding movie, which resulted in the retrieval of unmatched knowledge, and the final response was inconsistent with the context.

The possible reason for the above is that our time-aware mechanism assigns too little attention weight to the 3rd turn of dialog when generating the query for the 5th turn, as Our time-aware mechanism learns the importance of utterances according to the order in which they appear in the dialog session. Therefore, the query generation module did not understand ``the film" in the 5th turn of dialog, referring to the ``Lighting Up The Stars" in the 3rd turn. Some other errors and their analysis can be found in \ref{c}.
\section{Conclusion}
This paper proposes a knowledge-driven dialog system that can dynamically retrieve knowledge via query generation as the conversation progresses and generate knowledgeable responses. We propose a simple time-aware mechanism that performs well in our query generation module, and we demonstrate the effectiveness of our system in an actual application environment. Our proposed query and response generation modules outperform their respective comparison models. The human evaluation results also show that our dialog system can dynamically retrieve knowledge via query generation and generate informative responses in actual environments. Our work can inspire future researchers to think about ``where knowledge comes from". For future work, we think it is meaningful to reduce the difference between the query generation module and the response generation module or use the same module to generate queries and responses because it can significantly reduce the number of parameters of the whole system to more suitable for industrialization.
\appendix
\section{Impact of generated queries}\label{a}
\begin{figure}[htbp]
	\centering
	\includegraphics[width=1\columnwidth]{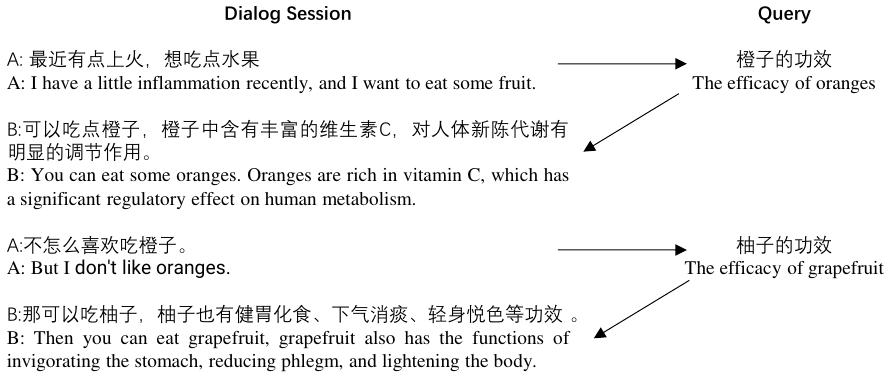}
	\caption{The impact of generated queries.}
	\label{figA1}
\end{figure}
\begin{table}[htbp]
	\centering
	\begin{tabular}{@{}cccc@{}}
		\toprule
		\textbf{Model Input}      & \textbf{Q\_F1} & \textbf{Q\_BLEU-1} & \textbf{Q\_BLEU-2}  \\ \midrule
		+generated queries                & 0.357       & 0.276           & 0.256        \\ 
		-generated queries & 0.327       & 0.228           & 0.214          \\\bottomrule
	\end{tabular}
	\caption{The experimental results for the impact of generated queries. Here, the model we used is the earlier version of our implementation, however, the conclusion drawn is general.}
	\label{tabA1}
\end{table}
Intuitively, the queries generated should have a particular prompting effect on the query to be generated. For example, as shown in Figure \ref{figA1}, if we want to generate a query such as "the effect of grapefruit," the previously generated query "the effect of oranges" will play a certain role as a hint. Moreover, the earlier implementation of our query generation module also proved our point, as illustrated in Table \ref{tabA1}. The metrics have been significantly improved if we concatenate the utterances in dialog history and their corresponding generated queries as the input of the encoder.
\section{Empty query and knowledge}\label{b}
During the collection of the Dusinc dataset, if the BOT does not use extra knowledge to generate the response, the query and knowledge provided in this dialog round will be ``empty". We do not perform any additional operations for the response generation module and directly use the dialogue history as the only input source to the encoder. In this way, our CPT-DoHA model degenerates into a CPT-Large model. However, for our query generation module, we filter out those samples with ``empty'' queries during training because using them as training samples will make the module seriously inclined to generate ``empty" queries. Moreover, in the human evaluation, we found that generating a query and retrieving knowledge can make the final response more informative for the vast majority of conversations. Therefore, we finally decided not to consider the case where the query is empty and generated a query for each round of dialogue. The results of human evaluation also show that our decision is correct.
\begin{figure}[!t]
	\centering
	\includegraphics[width=1\columnwidth]{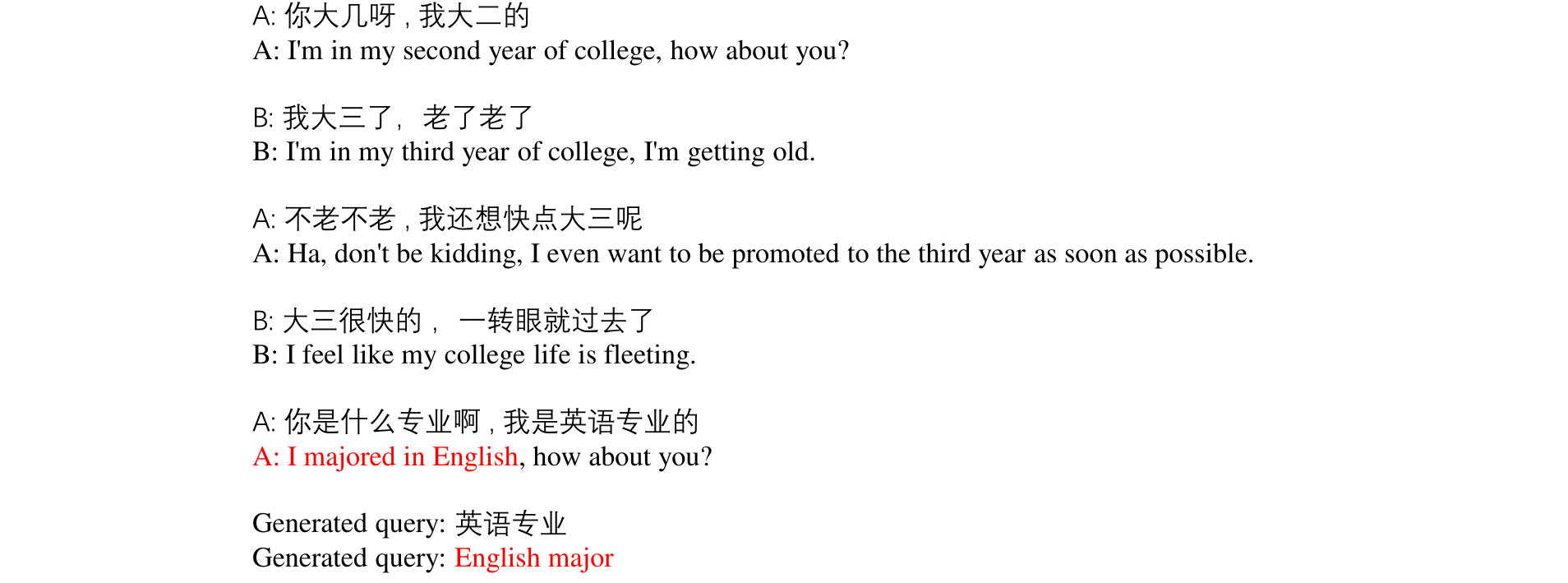}
	\caption{Reasoning error.}
	\label{figC1}
\end{figure}
\begin{figure}[!t]
	\centering
	\includegraphics[width=1\columnwidth]{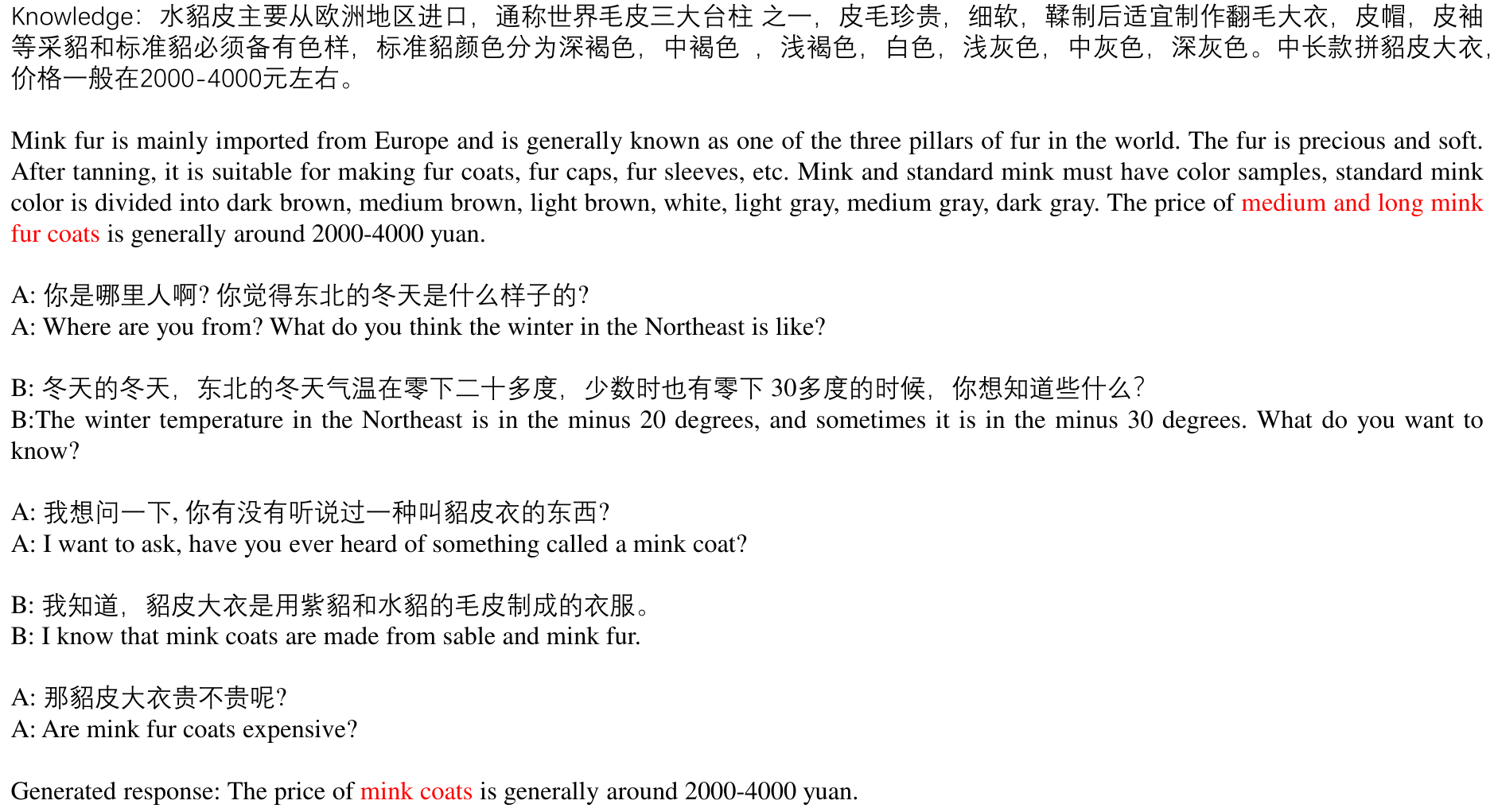}
	\caption{Consistency error.}
	\label{figC2}
\end{figure}
\section{Additional cases}\label{c}
We analyzed a lot of cases for our query generation module and response generation module independently. We found that for the query generation module, most of the errors are caused by the lack of reasoning ability of the model, such as Figure \ref{figC1}. When speaker A asks about speaker B's major, the module does not recognize that the ``English major'' is an attribute of the speaker A, so the query is wrongly generated. For the response generation module, the partial inconsistency of knowledge is the leading cause of errors, as shown in Figure \ref{figC2}, ``the price of medium and long mink fur coats is 2000-4000 yuan'', instead of ``the price of mink coats is 2000-4000 yuan''.
  \bibliographystyle{elsarticle-num} 
  \bibliography{cas-refs}

\par\noindent
\parbox[t]{\linewidth}{
	\noindent\parpic{\includegraphics[height=1.2in,width=1in,clip,keepaspectratio]{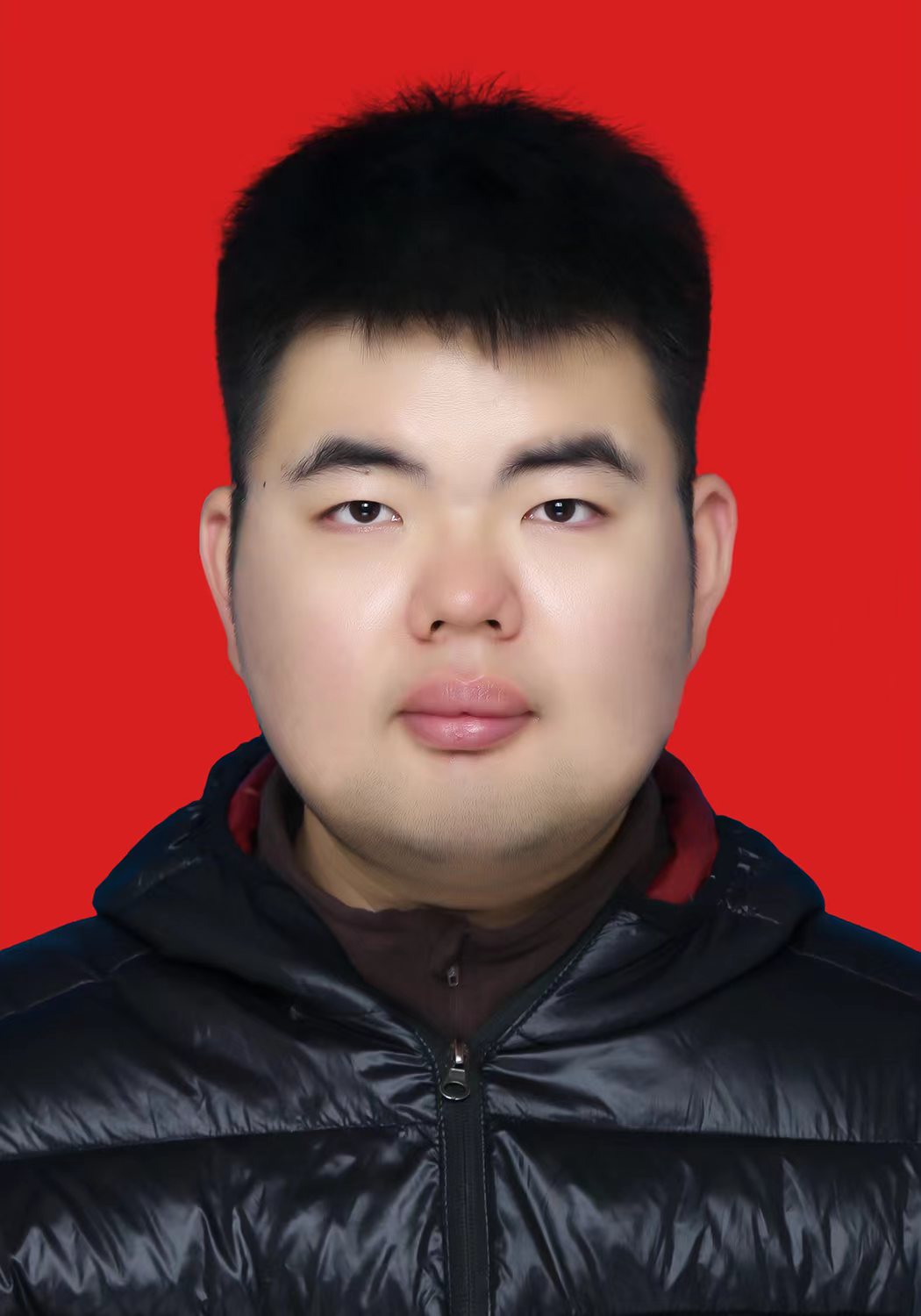}}
	\noindent {\bf Zhongtian Hu}\
	is going on pursuing a Ph.D. in Computer Science and Technology from Northwestern Polytechnical University. His research interests include Natural Language Processing, Question Answering system and Dialogue system.}
\vspace{4\baselineskip}
\par\noindent
\parbox[t]{\linewidth}{
	\noindent\parpic{\includegraphics[height=1.2in,width=1in,clip,keepaspectratio]{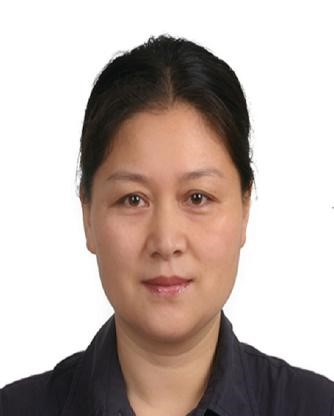}}
	\noindent {\bf Lifang Wang}\
	is a professor in the School of Computer Science, at Northwestern Polytechnical University in Xi'an, China. She received her Ph.D in Computer Science and Technology from Northwestern Polytechnical University. Her research interests include Electronic Commerce Technology, Internet \& Information Security, Cloud Computing, Cloud storage, Cloud security, Machine Learning, Deep Learning, Natural Language Processing.}
\vspace{4\baselineskip}
\par\noindent
\parbox[t]{\linewidth}{
	\noindent\parpic{\includegraphics[height=1.2in,width=1in,clip,keepaspectratio]{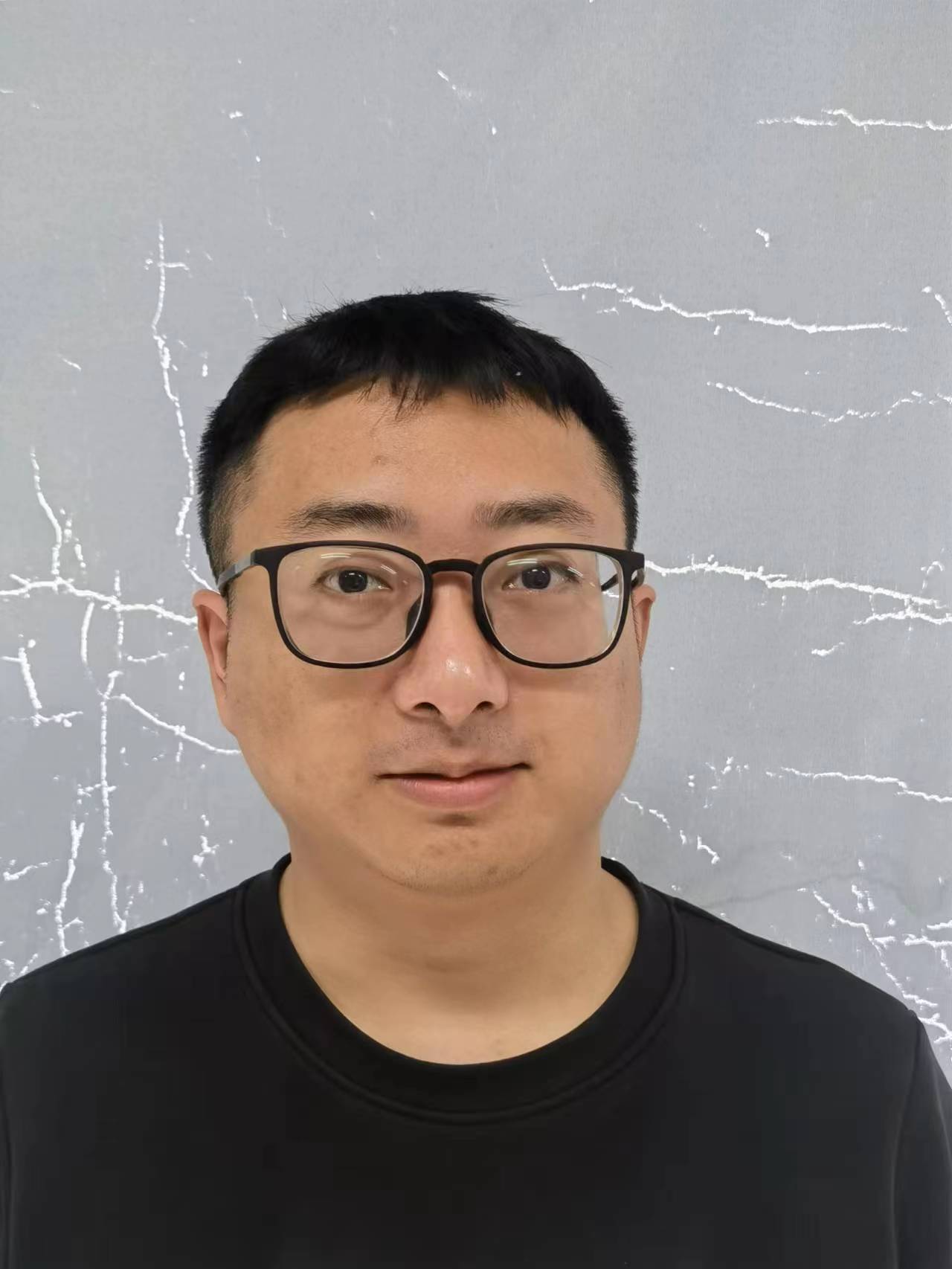}}
	\noindent {\bf Yangqi Chen}\
	is going on pursuing a M.sc. in Computer Science and Technology from Northwestern Polytechnical University. His research interests include Natural Language Processing, Question Answering system and Dialogue system.}
\vspace{4\baselineskip}
\par\noindent
\parbox[t]{\linewidth}{
	\noindent\parpic{\includegraphics[height=1.2in,width=1in,clip,keepaspectratio]{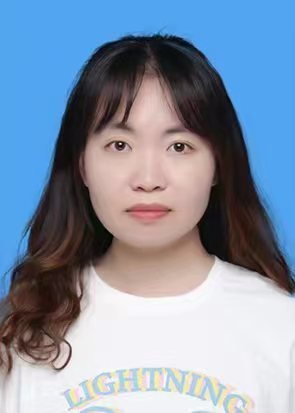}}
	\noindent {\bf Yushuang Liu}\
	is going on pursuing a M.sc. in Computer Science and Technology from Northwestern Polytechnical University. Her research interests include Natural Language Processing, Question Answering system and Dialogue system.}
\vspace{4\baselineskip}
\par\noindent
\parbox[t]{\linewidth}{
	\noindent\parpic{\includegraphics[height=1.2in,width=1in,clip,keepaspectratio]{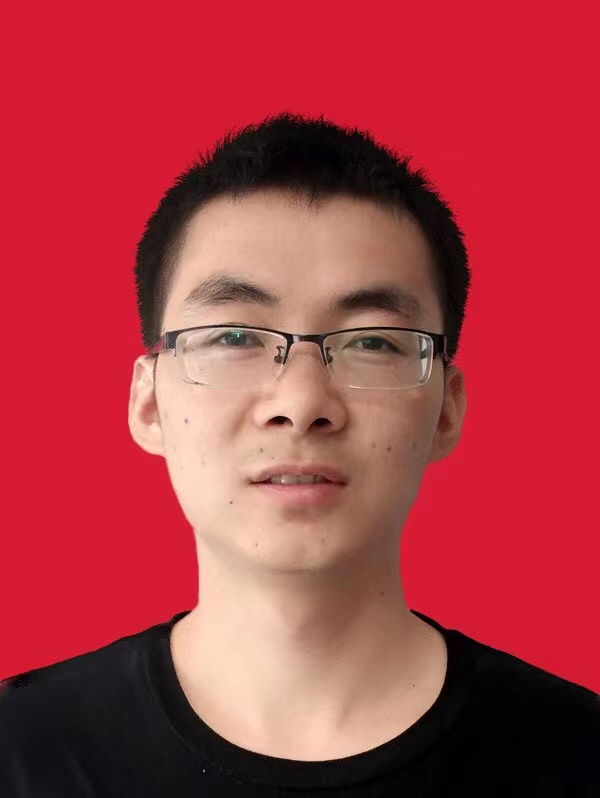}}
	\noindent {\bf Meng Zhao}\
	is a Ph.D. candidate at Northwestern Polytechnical University in Xi’an, China. He was born in Henan, China. He received his B.S degrees and M.S degrees in Computer Science and Technology in 2014, 2017 from Henan Normal University, Henan University of Technology respectively. His research interests include dialogue system and text generation.}
\vspace{4\baselineskip}
\par\noindent
\parbox[t]{\linewidth}{
	\noindent\parpic{\includegraphics[height=1.2in,width=1in,clip,keepaspectratio]{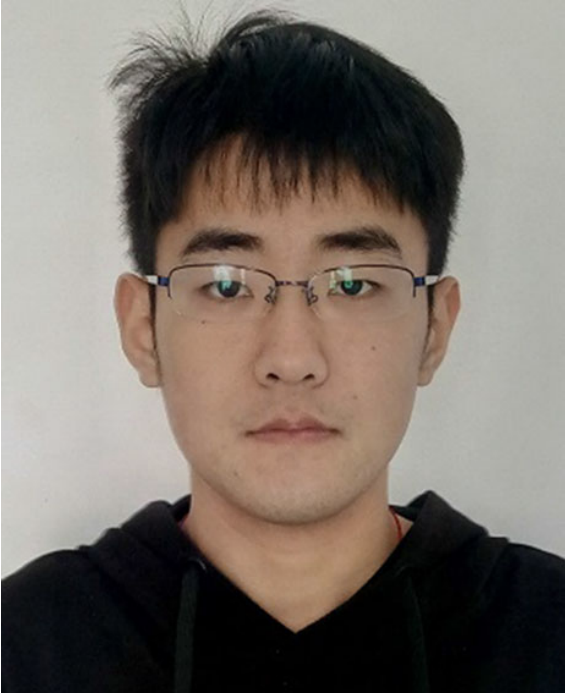}}
	\noindent {\bf Ronghan Li}\
	is a A.P. in Computer Science and Technology from Xidian University. He received his Ph.D. degrees in Computer Science and Technology in 2022 from Northwestern Polytechnical University. His research interests include Natural Language Processing, Question Answering system and Dialogue system.}
\vspace{4\baselineskip}
\par\noindent
\parbox[t]{\linewidth}{
	\noindent\parpic{\includegraphics[height=1.2in,width=1in,clip,keepaspectratio]{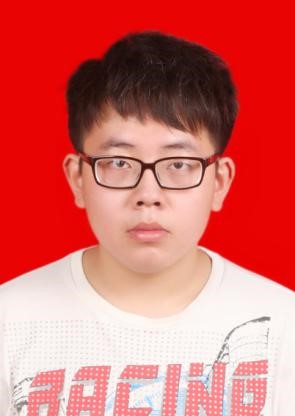}}
	\noindent {\bf Xinyu Lu}\
	is going on pursuing a Ph.D. in Computer Science and Technology from Northwestern Polytechnical University. His research interests include Natural Language Processing, Question Answering system and Knowledge Representation Learning.}
\vspace{4\baselineskip}
\par\noindent
\parbox[t]{\linewidth}{
	\noindent\parpic{\includegraphics[height=1.2in,width=1in,clip,keepaspectratio]{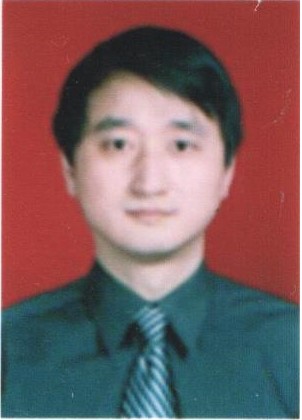}}
	\noindent {\bf Zejun Jiang}\
	is a professor in the School of Computer Science, at Northwestern Polytechnical University in Xi'an, China. He received his B.S degrees and M.S degrees in Computer Science and Technology in 1985, 1988 respectively from Northwestern Polytechnical University. His research interests include deduplication in distributed systems, Cloud Computing, Cloud storage, Deep Learning, Natural Language Processing.}
\vspace{4\baselineskip}
\end{document}